\documentclass[pmlr]{jmlr}

\usepackage{longtable}

 \usepackage{booktabs}
 
\usepackage[load-configurations=version-1]{siunitx} 

\usepackage{amsmath}
\usepackage{amssymb}
\usepackage{mathtools}
\usepackage{multirow}
\usepackage{outlines}
\usepackage[htt]{hyphenat}

\makeatletter
\def\set@curr@file#1{\def\@curr@file{#1}} 
\makeatother

\theorembodyfont{\upshape}
\theoremheaderfont{\scshape}
\theorempostheader{:}
\theoremsep{\newline}

\jmlrvolume{219}
\jmlryear{2023}
\jmlrworkshop{Machine Learning for Healthcare}

\title[RadGraph2]{RadGraph2: Modeling Disease Progression in Radiology Reports via Hierarchical Information Extraction}

\begin{document}

\author{\Name{Sameer Khanna*}
       \Email{sameer\_khanna@berkeley.edu}\\ 
       \addr Stanford University\\
       \AND
       \Name{Adam Dejl*}
       \Email{adamdejl@mit.edu}\\ 
       \addr Massachusetts Institute of Technology\\
       \AND
       \Name{Kibo Yoon}
       \Email{kiboyoonmd@gmail.com}\\ 
       \addr Hanyang University\\
       \AND
       \Name{Quoc Hung Truong}
       \Email{brain01@vinbrain.net}\\ 
       \addr Vinbrain\\
       \AND
       \Name{Hanh Duong}
       \Email{v.hanhduong@vinbrain.net}\\ 
       \addr Vinbrain\\
       \AND
       \Name{Agustina Saenz†}
       \Email{asaenz@bwh.harvard.edu}\\ 
       \addr Harvard Medical School\\
       \AND
       \Name{Pranav Rajpurkar†}
       \Email{pranav\_rajpurkar@hms.harvard.edu }\\ 
       \addr Harvard Medical School}

\def\thefootnote{*}\footnotetext{These authors contributed equally to this manuscript.}
\def\thefootnote{†}\footnotetext{These authors contributed equally to this manuscript.}

\maketitle

\begin{abstract}


We present RadGraph2, a novel dataset for extracting information from radiology reports that focuses on capturing changes in disease state and device placement over time. We introduce a hierarchical schema that organizes entities based on their relationships and show that using this hierarchy during training improves the performance of an information extraction model. Specifically, we propose a modification to the DyGIE++ framework, resulting in our model HGIE, which outperforms previous models in entity and relation extraction tasks. We demonstrate that RadGraph2 enables models to capture a wider variety of findings and perform better at relation extraction compared to those trained on the original RadGraph dataset. Our work provides the foundation for developing automated systems that can track disease progression over time and develop information extraction models that leverage the natural hierarchy of labels in the medical domain.



\end{abstract}

\begin{figure*}[!tb]
  \label{fig:RadGraph2-overview}
  \includegraphics[width=1.0\linewidth]{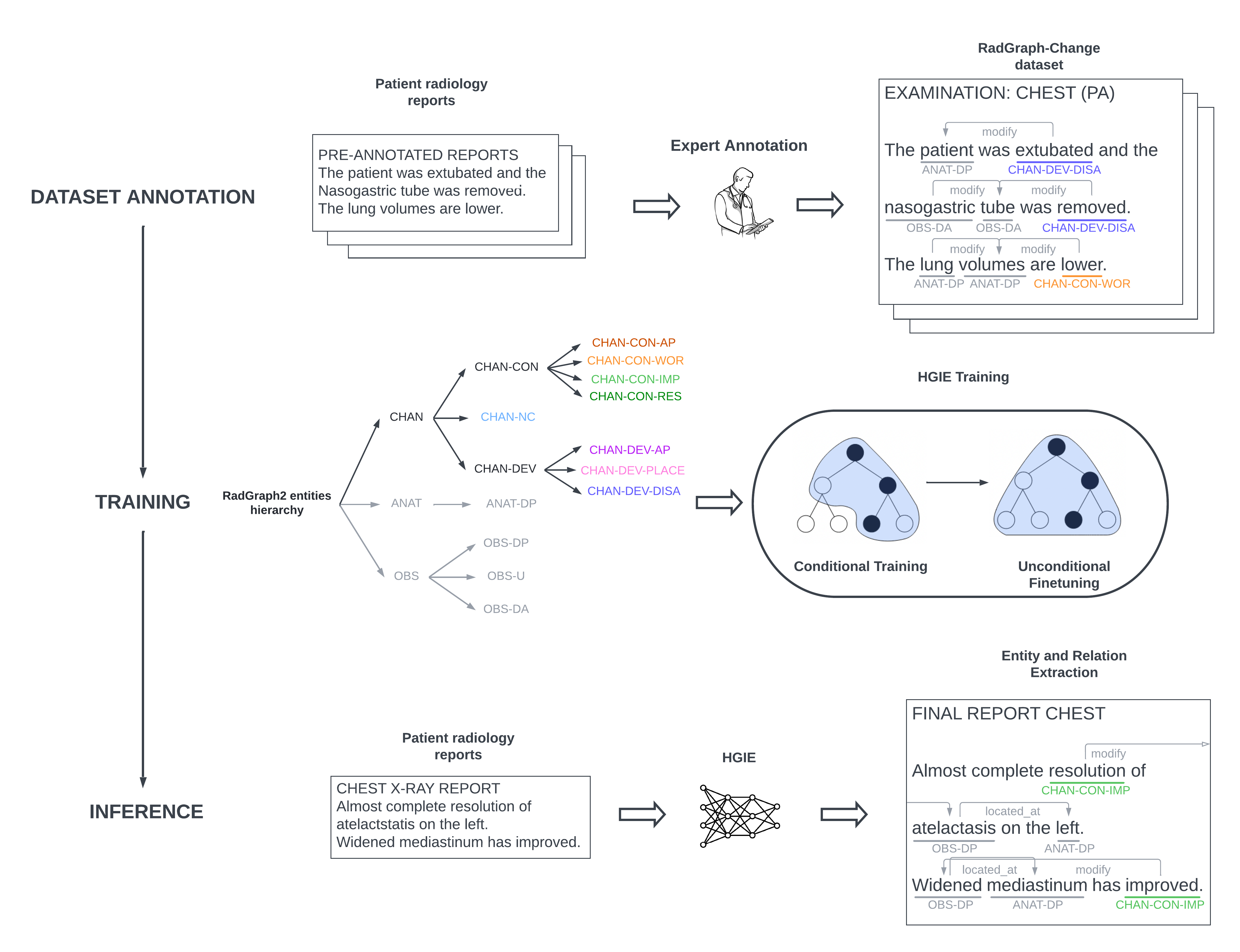}
  \caption{Overview of the HGIE training process and its application to a downstream task of change classification for radiology reports in the RadGraph2 dataset. Change entities, which have been added as part of our expanded information schema, are highlighted.}
\end{figure*}

\section{Introduction}
\label{sec:intro}


Extracting and understanding information held within medical reports has become a central task in medical artificial intelligence, fueled by the creation of various medical report datasets. Recently, a push has been made away from conventional disease classification labels and labeling methods that can only mark the presence or absence of a narrow set of diseases \citep{chexpert-irvin, chexpert++-mcdermott, chexbert-smit}. In its place, radiographs, graphs created from X-rays, are becoming more popular. They are capable of dynamically modeling rich contextual information by marking the level of presence and relations to other entities and modifications. They can also be used to support a wide range of clinical tasks. Recent work built upon the RadGraph dataset, the most widely-used medical report graph dataset, has shown that graphs built from medical reports are capable of facilitating a wide range of research in the fields of natural language processing, computer vision, and multi-modal learning in various medical domains \citep{chiu2022application, li2022cross, yang2022knowledge, mcgrath2022optimizing, wu2023medklip, stupp2022structured, li2023dynamic, Yu2022.08.30.22279318, jeong2023multimodal}.

Nonetheless, RadGraph is still far from encapsulating all the information a patient’s medical report contains; for example, it is missing key information regarding prior patient data. Like most past work on medical report interpretation, it is focused on detecting conditions in a single scan or instance. The schema utilized within the original dataset does not encapsulate the relations between various entities and only focuses on disease info rather than including information regarding disease progression and device placement or removal. One key component of clinical reports, such as those used in radiology, is comparison to prior patient data, which enables medical professionals to track disease progression over time. There has therefore been a strong push for AI tools that can not only interpret findings in a single study but can also integrate information from prior reports to provide a complete clinical picture \citep{acosta2022need, rameshcxr, moor2023foundation, rajpurkar2023current}.

In this work, we construct Radgraph2, a new dataset of 800 chest X-ray expert annotations and an additional 220,913 annotated via inference. Radgraph2 combines the schema from the previous version of RadGraph with additional entities focused on characterizing priors, like disease progression and device information, along with clinically relevant contextual information in radiology reports.

The new Radgraph2 schema introduces a taxonomy to the entities found within a chest X-ray report, taking advantage of their natural hierarchy. We show that capturing relationships between entities via a hierarchy can lead to increases in performance through our model HGIE, an information extraction model that modifies the DyGIE++ framework \citep{wadden2019entity}. HGIE uses a two-phase training methodology in order to learn the entity hierarchy, achieving superior performance to DyGIE++ despite sharing the same model architecture.

In summary, our contributions are as follows:

\begin{itemize}
  \item We devise a novel schema for capturing change along with clinically relevant contextual information in radiology reports.

  \item We introduce a hierarchy to the schema, organizing entities based on their natural relationships. We show that taking advantage of this inherent organization can lead to performance improvements.
  
  \item We extend the RadGraph dataset with additional expert annotations, bringing the total number of X-ray reports densely annotated according to our schema to 800. This dataset is expanded via inference to include an additional 220,463 reports from MIMIC-CXR and 450 from CheXpert.

\end{itemize}

\subsection*{Generalizable Insights about Machine Learning in the Context of Healthcare}
The automatic identification of changes in a patient's condition from clinical notes is a critical task in healthcare, as information about past disease progression is vital for medical professionals. In this paper, we propose a hierarchical schema for modeling changes in radiology reports. We construct a new dataset for this schema and introduce a model trained to automatically identify changes based on the schema's hierarchy. Our approach enables automated systems to reason about the differences observed across multiple radiological studies, which may facilitate a variety of useful clinical applications, such as modeling patient healthcare trajectories and training large-scale medical imaging models. We also believe that our schema and general approach could be extended to other domains beyond radiology.

\section{Related Work}
\label{sec:related-work}

In this work, we aim to extract key entities and their relations from radiology reports, building upon the work done by RadGraph. Various other natural language processing (NLP) approaches have been developed and used to extract information from medical reports and radiology reports \citep{chexpert-irvin, peng2018negbio, chexpert++-mcdermott, chexbert-smit, jain2021visualchexbert}. The most popular approach uses automated radiology report labelers to label the reports within large-scale chest radiograph datasets like MIMIC-CXR and CheXpert \citep{johnson2019mimic, chexpert-irvin} for a variety of common medical conditions. While these labels capture information about the presence of particular diseases and conditions, they do not capture more rich contextual information, such as the relations between the concepts mentioned in each report nor do they capture vital information regarding the presence of priors.

\paragraph{Information Extraction in the Medical Sphere}

RadGraph2 utilizes entity and relation schemas to extract detailed information from radiology reports. While there are approaches proposed that aim to achieve the same thing using entity extraction schemas \citep{hassanpour2016information, sugimoto2021extracting} fact-focused schemas, \citep{steinkamp2019toward}, or even spatial relations \citep{datta2020understanding, datta2020rad}, such approaches are limited due to their requirement for task-specific datasets that are densely annotated by domain experts.

We build upon RadGraph, which extracts medical information as disjoint radiograph representations via a combination of physician-annotated reports and entity and relation extraction model architecture \citep{radgraph-jain}. This is in contrast to recent approaches like Chest ImaGenome, which structured chest X-ray reports as scene graphs using a joint rule-based approach that takes advantage of natural language processing (NLP) and an atlas-based bounding box detection pipeline \citep{wu2021chest}.

Our newly proposed schema enables the classification of changes in radiology reports, an endeavor that has been considered by other approaches recently. \citet{change-characterization-hassanpour} proposed an NLP system for characterizing several common types of changes, but their method largely relies on a set of pre-determined rules and does not allow for more fine-grained classification of changes and their context. \citet{nodular-change-classification-yuan} devised a deep learning architecture for the classification of changes in pulmonary nodular findings described in radiology reports, using a Siamese network to overcome data sparsity. However, their method conflates certain key change categories (e.g. new and indeterminate) and is also limited to sentence-level classification. \citet{gliomas-change-detection-noto} applied a random forest classifier to detect changes in radiology reports of patients with high-grade gliomas. However, their approach only considers two types of changes (stable and unstable) and operates on the report level, resulting in coarse-grained labels.

\paragraph{Studies on the Relationships Between Entities}
Our modified information extraction model, HGIE, is built upon the DyGIE++ framework, which has shown great success in extracting information from medical reports \citep{wadden2019entity, radgraph-jain}. The DyGIE++ framework is a unified, multi-task framework that supports multiple information extraction tasks including named entity recognition and relation extraction. HGIE attempts to build upon this framework to improve performance on both tasks by taking advantage of the taxonomy that defines the relationships between various entities.

While HGIE takes advantage of a fixed hierarchical structure to extract entities and relations, other methodologies have considered relationships between entities for information extraction in different manners. Cotype \citep{ren2017cotype} jointly embeds entity mentions, relation mentions, text features, and type labels into two low-dimensional spaces, where objects whose types are close will also have similar representations. Unlike HGIE, this approach autogenerates a noisy entity-type hierarchy for information extraction that is highly dependent on a large amount of labeled data. Fine-grained Entity Type Classification \citep{xu2018neural} adds a penalty term to the loss function that penalizes terms based on ancestor similarity. It is only capable of entity extraction, unlike HGIE which supports joint entity and relation extraction.

\section{RadGraph2 Schema}
\label{sec:schema}

We develop a novel hierarchical schema for entities and relations, which extends on the original RadGraph schema \citep{radgraph-jain}, to capture detailed information about changes and their context described in radiology notes. In its original formulation, the schema was designed to maximize the coverage and retention of clinically relevant information contained in the reports while remaining sufficiently simple to enable quick and reliable labeling. Our extended schema adheres to these design principles while introducing additional entity types to represent various kinds of changes.

We utilized an iterative approach to schema development that ensures we create accurate and reliable models for capturing different change types. In each iteration, we identified entities that represent specific change types, and we labeled reports accordingly using the latest version of the schema. To ensure the schema is effective, we collected feedback from medical practitioners on areas where the schema may be ambiguous or unable to accurately capture information from annotated notes. We paid particular attention to these cases to refine and improve the schema. This iterative process was continued until we were satisfied with the schema's coverage, faithfulness, and reliability.

\subsection{Entities}
Entities are objects associating contiguous spans of tokens (i.e. words and punctuation marks) with their corresponding entity types. In our schema, we retain all the Anatomy and Observation entity types from the original RadGraph dataset \citep{radgraph-jain}, adding several new entity types for describing changes. \texttt{CHAN-CON} entities (\texttt{CHAN-CON-AP} — condition appearance, \texttt{CHAN-CON-WOR} — condition worsening, \texttt{CHAN-CON-IMP} — condition improvement, and \texttt{CHAN-CON-RES} — condition resolution) refer to changes in medical conditions, while \texttt{CHAN-DEV} (\texttt{CHAN-DEV-AP} — device appearance, \texttt{CHAN-DEV-PLACE} — change in device placement and \texttt{CHAN-DEV-DISA} — device disappearance) refer to changes related to supporting devices and tubes used by the patient. Finally, \texttt{CHAN-NC} indicates there have been no changes in comparison with the prior studies. We explain the different entity types in more detail in the following section.

\subsubsection{Entity Types Descriptions \label{sec:entitydesc}}

The Anatomy entities (\texttt{ANAT-DP}) mark mentions of anatomical locations or body parts within the report, as well as modifiers of these mentions. For example, ``\textit{left lower lung}'' may be annotated as three \texttt{ANAT-DP} entities, one for each token.

The Observation entities (\texttt{OBS-DP}, \texttt{OBS-U} and \texttt{OBS-DA}) are used for tokens describing general impressions detailed in the given radiology note. These can include mentions of non-anatomical features visualized in the radiograph or descriptions of possible diagnoses. For instance, all separate tokens of the texts ``\textit{small pneumothorax}'', ``\textit{support devices}'', and ``\textit{within normal limits}'' could be marked as Observation entities. Three different observation types are used for indicating the uncertainty level associated with the given entity: Definitely Present (\texttt{OBS-DP}), Uncertain (\texttt{OBS-U}), and Definitely Absent (\texttt{OBS-DA}). For example, in a report stating that there is ``\textit{no evidence of pneumothorax}'', \textit{pneumothorax} could be annotated as an \texttt{OBS-DA} entity due to the assertion of the absence of appreciable pneumothorax.

In addition to the Anatomy and Observation entities, our schema also defines a set of Change entities (\texttt{CHAN-NC}, \texttt{CHAN-CON-AP}, \texttt{CHAN-CON-WOR}, \texttt{CHAN-CON-IMP}, \texttt{CHAN-CON-RES}, \texttt{CHAN-DEV-AP}, \texttt{CHAN-DEV-PLACE}, \texttt{CHAN-DEV-DISA}) describing change or lack of change in comparison with the prior radiological studies.

\begin{itemize}

\item Change -- No Change (\texttt{CHAN-NC}) entities denote that a certain condition or observation remained the same since a previously performed examination. For instance, in the texts ``\textit{no change in right pleural effusion}'' and ``\textit{bibasilar atelectasis persists}'', \textit{change} and \textit{persists} could be marked as \texttt{CHAN-NC}.

\item Change in Condition (\texttt{CHAN-CON}) entities mark various kinds of changes in the observed medical conditions. We utilize four distinct sub-types for these entities: Condition Appearance (\texttt{CHAN-CON-AP}), Condition Worsening (\texttt{CHAN-CON-WOR}), Condition Improvement (\texttt{CHAN-CON-IMP}) and Condition Resolution (\texttt{CHAN-CON-RES}). \texttt{CHAN-CON-AP} indicates that a new adverse medical condition has been observed in the given patient. As an example, \textit{new} in ``\textit{new pulmonary edema}'' could be labeled with this entity type. \texttt{CHAN-CON-WOR} marks worsening in a certain aspect of the patient's clinical state. For example, a token \textit{increased} in ``\textit{pleural effusion has increased}'' would be appropriately marked as \texttt{CHAN-CON-WOR}. \texttt{CHAN-CON-IMP} indicates a general improvement, e.g. \textit{diminished} in ``\textit{pleural effusion has diminished slightly}''. Finally, \texttt{CHAN-CON-RES} signifies a complete resolution of a particular medical condition. Thus, in a note text such as ``\textit{opacity has completely cleared}'', the word \textit{cleared} might be assigned a \texttt{CHAN-CON-RES} label·

\end{itemize}

Our schema also defines three entity types for representing changes associated with medical devices: Device Appearance (\texttt{CHAN-DEV-AP}), Change in Device Placement (\texttt{CHAN-DEV-PLACE}), and Device Disappearance (\texttt{CHAN-DEV-DISA}). \texttt{CHAN-DEV-AP} entities denote that the patient has been fitted with a new medical device or tool; for example \textit{intubated} in the report commentary ``\textit{patient has been inubated}''. \texttt{CHAN-DEV-PLACE} entity type describes changes in the position of a medical device compared to previous studies. For instance, \textit{migrated} in ``\textit{NG tube has migrated proximally}'' could be marked as a \texttt{CHAN-DEV-PLACE} entity. Finally, c{CHAN-DEV-DISA} indicates that a medical device or tool was detached or removed from the patient, as in ``\textit{nasogastric tube was removed}'' where \textit{removed} can be labeled with this entity type.

It is important to note the natural hierarchy within the entity types in our annotation schema. Anatomy (\texttt{ANAT}), Observation (\texttt{OBS}) and Change (\texttt{CHAN}) are more general entity types. These can be further subdivided into even more fine-grained categories. For example, a change (\texttt{CHAN}) can be further described as an improvement in condition (\texttt{CHAN-CON-IMP}). We utilize this structure of the entity labels in our hierarchical model, described in section \ref{sec:hierarchical}. The full taxonomy of our entity types is visualized in Figure 1.

\subsection{Relations}
Relations in our schema are defined as directed edges between entities. Similarly to entities, each relation is associated with its corresponding label. In our work, we utilize the same set of relation labels as in \cite{radgraph-jain} with slight modifications to their definitions. There are three types of relations overall: Modify (\texttt{modify}), Located At (\texttt{located\_at}), and Suggestive Of (\texttt{suggestive\_of}).

\subsubsection{Relation Types Descriptions \label{sec:reldesc}}

The \texttt{modify} relation is used for associating Observation and Anatomy modifiers with their main entity, as well as for connecting Change entities to Observation and Anatomy entities the change relates to. For example, in the report text samples ``\textit{left lower lung}'' and ``\textit{nasogastric tube was removed}'' that we introduced earlier, the entity pairs (\textit{left}, \textit{lung}), (\textit{lower}, \textit{lung}), (\textit{nasogastric}, \textit{tube}) and (\textit{removed}, \textit{tube}) would be connected by a \texttt{modify} relationship.

The \texttt{located\_at} relationship links Observation and Anatomy entities and indicates that the source Observation is related to the target Anatomy. While it commonly describes the location at which a certain observation was noted (hence the name of this relation type), it can also be used to describe other kinds of relationships between Observation and Anatomy entities. For instance, in the note text ``\textit{cardiac silhouette is not enlarged}'', the \texttt{OBS-DA} entity \textit{enlarged} and the \texttt{ANAT-DP} entity \textit{silhouette} could be connected via a \texttt{located\_at} relation.

Finally, the \texttt{suggestive\_of} relation represents cases in which a presence of a certain Change or Observation is derived from another Change or Observation. For example, in the report sentence ``\textit{The lungs are hyperinflated suggestive of COPD}.'', the entity pair (\textit{hyperinflated}, \textit{COPD}) could be in a \texttt{suggestive\_of} relationship.

\section{RadGraph2 Dataset}
The RadGraph2 dataset is composed of 800 chest X-ray reports annotated according to a hierarchical schema. It adds fine-grained information about changes in comparison to priors while also adding 200 additional labeled reports, considerably increasing the sizes of the training and test sets.

\subsection{Annotation Process}
For the re-annotation of the reports from the original RadGraph dataset \citep{radgraph-jain} according to our schema, we liaised with a team of four board-certified radiologists and one academic hospitalist. The 600 reports from the RadGraph development and test datasets were imported into a specialized text labeling platform Datasaur \citep{datasaur} and split among the available annotators. The annotators were instructed to focus on correctly marking the mentions of change and the relations associated with these mentions while making minimal modifications to the entities and relations not associated with any changes, as we considered those to have been labeled with sufficient quality. Nevertheless, the annotators were free to correct blatant mistakes in any aspect of the annotations. Annotation instructions can be found in the supplementary material.

We conducted a pilot labeling task where they were provided with a set of 15 reports to label according to our instructions. During the pilot, we evaluated the agreement between the annotators as well as the general reliability of the labeling. We also took this opportunity to provide personalized feedback to each of the annotators so as to rectify possible misunderstandings of the labeling instructions. As the pilot study was used for training purposes, we discarded the labels obtained during the pilot task in order to maintain the integrity of the dataset.

In addition to the 600 reports from the original RadGraph development and test sets, we also extended RadGraph2 with 200 additional reports randomly sampled from the MIMIC-CXR portion of the RadGraph inference dataset. To speed up and simplify the task, we based the initial labels for these reports on the output of the RadGraph Benchmark model included in the inference set. The annotators were instructed to label the entities and relations associated with changes, as well as to correct any possible mistakes or deficiencies in the entities and relations identified by the benchmark model.

We achieved high agreement during the annotation process. The minimum pairwise agreement, computed using Cohen's Kappa, was 0.9943, and the median was 0.9963.

\subsection{Data Overview}
\label{section:dataoverview}
The resulting RadGraph2 dataset is comprised of 800 densely annotated reports with 23457 entities and 17373 relations. It is split into three partitions: a train partition with 575 annotated reports originating from MIMIC-CXR, a validation partition with 75 MIMIC-CXR reports, and a test partition with 100 reports from MIMIC-CXR and 50 reports from CheXpert. We give the detailed label statistics for the different dataset partitions in Table \ref{tab:label-statistics}.

\begin{table*}[!t]
    \small
     \caption{Label statistics for the RadGraph2 datasets}
     \label{tab:label-statistics}
     \centering
     \begin{center}
     \resizebox{0.75\textwidth}{!}{%
     \begin{tabular}{ c c c c c } 
        \toprule
        & \textbf{Train (\%)} & \textbf{Validation (\%)} & \textbf{MIMIC-CXR Test (\%)} & \textbf{CheXpert Test (\%)} \\ 
        \toprule
        \texttt{ANAT} & 7081 (41.9) & 987 (43.7) & 1186 (42.6) & 641 (42.7)\\
        \texttt{OBS-DP} & 5822 (34.4) & 731 (32.3) & 901 (32.4) & 543 (36.2)\\
        \texttt{OBS-U} & 659 (3.9) & 87 (3.8) & 114 (4.1) & 47 (3.1)\\
        \texttt{OBS-DA} & 1957 (11.6) & 276 (12.2) & 424 (15.2) & 167 (11.1)\\
        \texttt{CHAN-NC} & 814 (4.8) & 102 (4.5) & 99 (3.6) & 69 (4.6)\\
        \texttt{CHAN-CON-AP} & 70 (0.4) & 5 (0.2) & 4 (0.1) & 5 (0.3)\\
        \texttt{CHAN-CON-WOR} & 186 (1.1) & 35 (1.5) & 14 (0.5) & 17 (1.1)\\
        \texttt{CHAN-CON-IMP} & 139 (0.8) & 18 (0.8) & 17 (0.6) & 5 (0.3) \\
        \texttt{CHAN-CON-RES} & 23 (0.1) & 4 (0.2) & 1 (0.0) & 0 (0.0) \\
        \texttt{CHAN-DEV-AP} & 21 (0.1) & 3 (0.1) & 4 (0.1) & 5 (0.3) \\
        \texttt{CHAN-DEV-PLACE} & 19 (0.1) & 1 (0.0) & 1 (0.0) & 1 (0.1) \\
        \texttt{CHAN-DEV-DISA} & 44 (0.3) & 7 (0.3) & 3 (0.1) & 5 (0.3) \\
        \midrule
        Total Entities & 16913 (100.0) & 2260 (100.0) & 2783 (100.0) & 1501 (100.0) \\
        \midrule
        \texttt{modify} & 7945 (63.4) & 1085 (64.1) & 1209 (61.0) & 765 (65.6) \\
        \texttt{located\_at} & 4088 (32.6) & 536 (31.7) & 694 (35.0) & 336 (28.8) \\
        \texttt{suggestive\_of} & 500 (4.0) & 71 (4.2) & 78 (3.9) & 66 (5.7) \\
        \midrule
        Total Relations & 12533 (100.0) & 1692 (100.0) & 1981 (100.0) & 1167 (100.0) \\
        \bottomrule
     \end{tabular}%
     }
     \end{center}
\end{table*}

\section{HGIE}



\label{sec:hierarchical}
Unlike traditional text-to-graph models like DyGIE++ where entities are not considered to have a type hierarchy, our proposed model HGIE aims to take advantage of the inherently structured organization of our labels to improve information extraction performance.

\paragraph{Hierarchical recognition}
Our hierarchical recognition (HR) system used in HGIE utilizes an entity taxonomy. We note that there are inherent relationships between the various entities used to label our graphs. For example, \texttt{CHAN-CON-WOR} and \texttt{CHAN-CON-AP} are both entities that refer to changes in a patient's condition. Taking advantage of these relationships between the various entities, we construct the entity taxonomy shown in Figure 1.

We use a BERT-based model as our backbone for the HR task, with the objective of extracting 12 scalar outputs. Each output is assumed to represent the conditional probability of an entity being true given that its parent in the entity hierarchy is true. During the inference phase, however, all the entities should be unconditionally predicted. Note that under such a training regimen, a trained model's unconditional probabilities can be calculated from the output using the probability chain rule via a simple application of the Bayes rule. 

For example, the probability for a given leaf node can be computed as shown in Equation 1. Here, $N_n$ denotes the leaf node. $N_0$ denotes the root node. The root node is defined as a dummy node that all entity types inherit from. Thus, no matter what the given true entity type is, the root node is always a correct ancestor; in other words $\mathrm {P}(N_0) = 1$. $N_k$ for $k$ from 0 to the tree's max depth are defined such that $N_{k}$ is a child of $N_{k-1}$.

\begin{equation}
\begin{aligned}
    \mathrm{P}(N_n) = \mathrm {P}(N_0) * \prod_{k=1}^{n} \mathrm {P}(N_k | N_{k-1})
\end{aligned}
\end{equation}

Here, we condition the probability of a given entity type being correct on the probability that all of its ancestors are correct as well. This ensures that the predicted unconditional probability of a parent entity is guaranteed to be greater than or equal to its children's entities. Our model is trained in two phases, first conditionally via a fine-to-coarse tree-based loss function before being finetuned unconditionally.

\paragraph{First Phase: Initialization under Conditional Training}

In the first phase of entity training, the HR system is trained on data under the condition that its parent class is positive. This follows what has been done in other work on hierarchical classification \citep{redmon2017yolo9000, roy2020tree, yan2015hd, chen2019deep, pham2021interpreting}. The intention behind this training regime is that it directly models the conditional probabilities of the entities by learning the dependent relationships between parent and child entities and concentrating on distinguishing lower-level labels, in particular the leaf entities.

\begin{figure}[!tb]
  \begin{center}
  \includegraphics[width=0.5\linewidth]{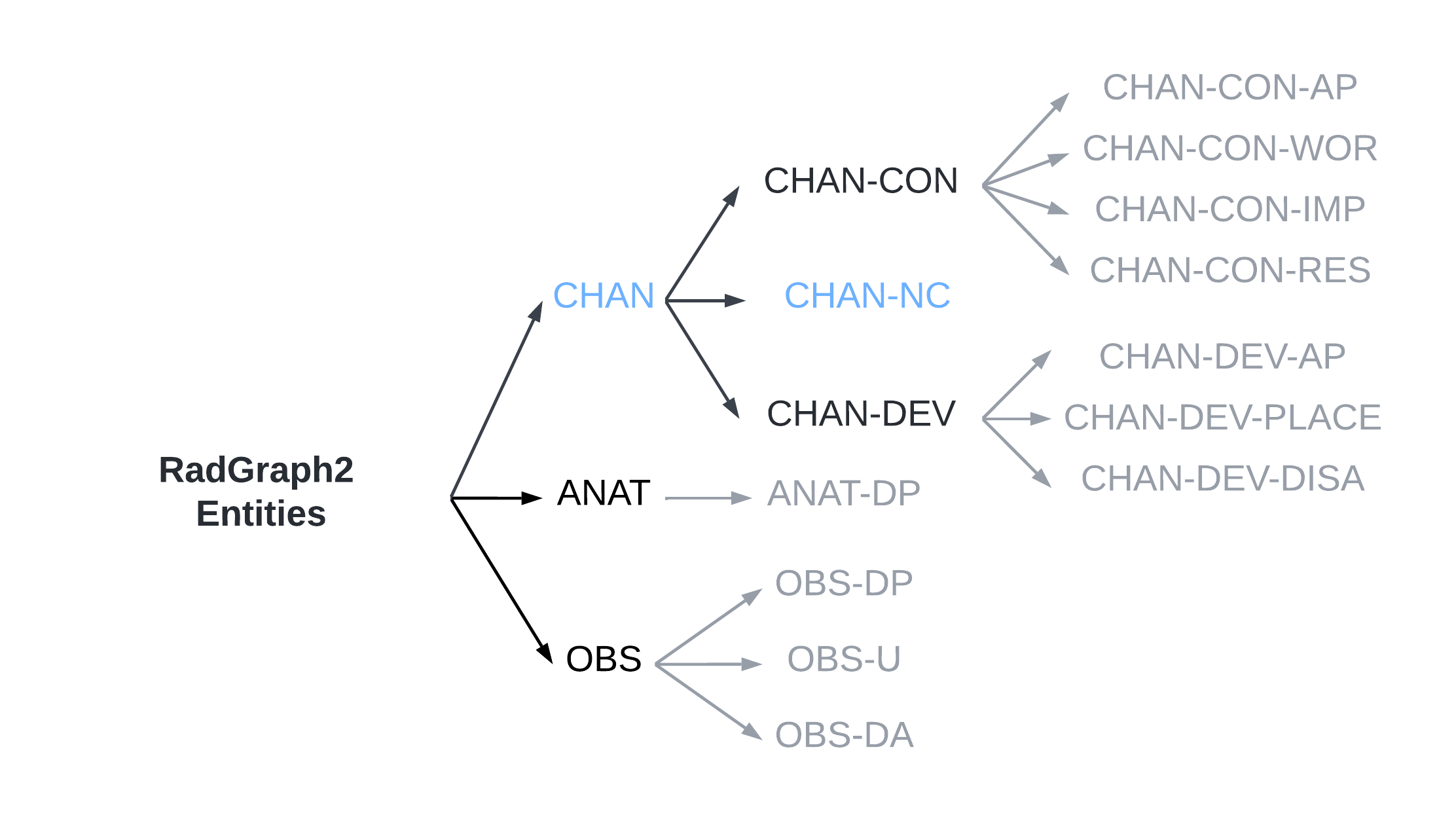}
  \caption{Visualization of Conditional Training. Here, the correct label is \texttt{CHAN-NC}. Black refers to negative labels, blue refers to positive labels. Under conditional training, we only penalize at the level of difference. Greyed-out portions of the tree are not part of the activation area in the loss function. In this example, the correct type is CHAN-NC, so we would penalize at the ANAT/OBS level, not at the ANAT-DP/OBS-DP/OBS-U/OBS-DA level. This improves computational complexity while also not hurting performance, as the probability of ANAT and OBS encompasses the probabilities of its children nodes. Additionally, we don't penalize or consider the nodes at levels beyond the correct classification of the node, in this example the CHAN-CON-AP/CHAN-CON-WOR/CHAN-CON-IMP/CHAN-CON-RES/CHAN-DEV-AP/CHAN-DEV-PLACE/CHAN-DEV-DISA. This is because their probabilities are already handled by comparing against CHAN-CON and CHAN-DEV.}
  \end{center}
\end{figure}

For each entity, the losses are only calculated on entities whose parent entity is also positive. Figure 2 shows an example of how this works using the RadGraph2 entity hierarchy.

Training under a conditional probability regimen could improve the model's initialization as it attempts to focus the model on identifying entities under the same parent entity, rather than having to discriminate the entity across all other possible labels. This can help to ease confounding and convergence factors. 

The idea behind the regimen is that it can help alleviate the issue of low label prevalence due to lower reliance on negative samples that occurs within the medical report problem space.

\paragraph{Fine-to-Coarse Depth-Dependant Hierarchical Loss Function}
For conditional training, we compute a sum of entity recognition losses over each level within the entity tree. The entity tree is defined via $(N, E)$, where $N = \{ n_1, n_2, n_3, .., n_n \}$ are the set of nodes, and where $E \subseteq N \times N$ is the set of edges in the tree such that $(n_i, n_j) \in E$ if and only if $n_i$ is the parent node of $n_j$. Without loss of generalizability, we assume that the first $k$ nodes correspond to the leaves of the tree, such that $\nexists n_j \in N$ where $(n_i, n_j) \in E$ for $i = 1, ..., k$. These nodes will refer to the finest-scaled entities in our system. This also means that if $\exists n_j \in N$ where $(n_i, n_j) \in E$, then $n_i$ is not a leaf node and is instead a more general entity definition of $n_j$. The label $l$ used for HR will be at the leaf level, that is $l \in \{1, ..., k\}$.

The depth of node $n_i$ under this definition of a tree is denoted via $\text{depth}(n_i)$, which refers to the number of edges between node $n_i$ and the root node. Thus, the max depth ($\text{d}_{max}$) is determined by the largest number of edges between any node and the root; in other words $\text{d}_{max} = \max_i \text{depth}(n_i)$.

The HR model outputs one logit, which we denote via $x_i$ per leaf node $n_i$. Thus, the probability $p$ associated with leaf node $n_i$ can be computed by applying the softmax function $\sigma$ on the logit $x_i$. Thus, $p_i$ denotes the i-th entry of the vector $\sigma(x)$. As discussed above, the probability of a parent node is defined to be the sum of the probabilities of its leaf nodes, leading the probability computation to be fine-to-coarse. Combining these principles together provides us with the formula for computing the probability for any node in the tree as shown in Equation 2.

\begin{equation}
    p_i = \begin{cases}
    \sigma(x_i) & \text{IF } 1 \leq i \leq k \\
    \sum\limits_{(n_i, n_j) \in E} p_j & \text{Otherwise}
    \end{cases}
\end{equation}

As we take the softmax of the logits, and all parent nodes probabilities are defined to be the summation of a subset of leaf nodes, $0 \leq p_i \leq 1$ for any node $n_i$, making this definition of the conditional probability for any node valid.

The correct label index $c$ provided to the HR model will be such that corresponds to the index of a leaf node ($c \in \{1, ..., k\}$), enabling easy loss computation for leaf nodes. In order to compute losses at all depths, we need to determine the correct node at depth $d$ which we denote $n_{cd}$. This computation is shown in Equation 3.

\begin{equation}
    n_{cd} = \begin{cases}
    n_c & \text{IF } \text{depth}(n_c) = d \\
    n_i & \text{IF } d < \text{depth}(n_c) \text{ AND } d = \text{depth}(n_i) \text{ AND } \text{anc}(n_i, n_c) \\
    \text{N/A} & \text{Otherwise}
    \end{cases}
\end{equation}

Here, $\text{anc}(n_i, n_c)$ is true if and only if node $n_i$ is an ancestor of node $n_c$; that is node $n_i$ represents a superclass of node $n_c$. Additionally, as mentioned in Section \ref{sec:hierarchical}, losses for conditional training are not computed for all nodes. Those nodes where the loss is not computed do not have a correct node, thus $n_{cd}$ is defined as N/A in those scenarios.

Putting these definitions together, the loss at depth $d$ is computed using the negative log loss of the probability at the given depth. The hierarchical loss $L$ is the summation of these losses at all depths in the tree as shown in Equation \ref{eq:hierarchical_loss}.

\begin{equation}
\label{eq:hierarchical_loss}
L = - \sum_{d = 0}^{\text{d}_{max}} \log(p_{n_{cd}})
\end{equation}

\paragraph{Second Phase: Fine-tuning under Unconditional Training}

\begin{figure}[!tb]
  \begin{center}
  \includegraphics[width=0.5\linewidth]{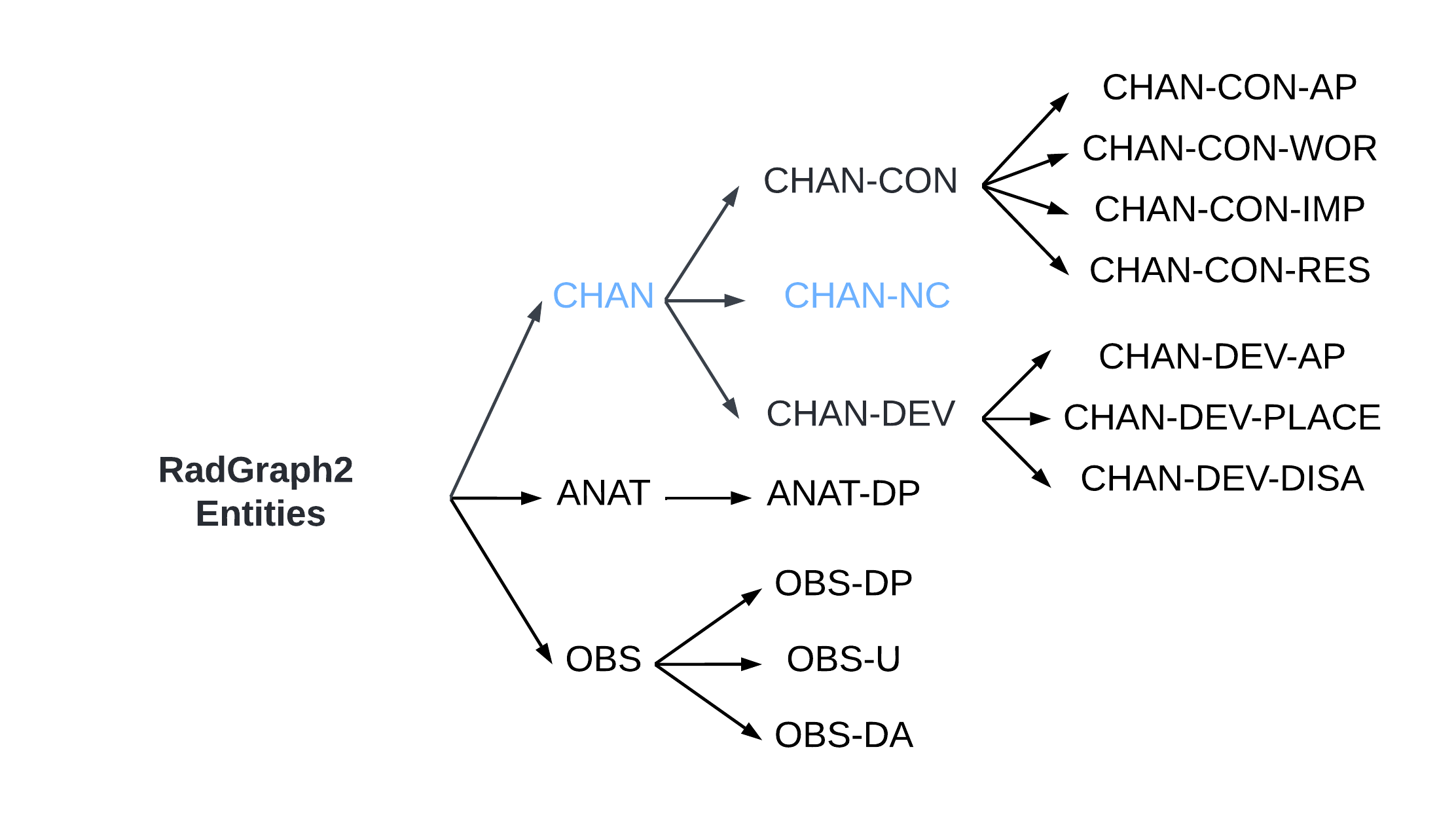}
  \caption{Visualization of Unconditional Fine-tuning. Here, the correct label is \texttt{CHAN-NC}. Black refers to negative labels, blue refers to positive labels. As the whole tree is part of the activation area in this phase, there are no greyed-out parts.}
  \end{center}
\end{figure}

This stage aims at improving the accuracy of unconditional probability predictions, which is used during inference and is thus critical to classification performance. To achieve this, we finetune our hierarchically trained network on the full dataset using a standard categorical cross-entropy loss function and smaller learning rates. This training stage aims at improving the capacity of the network in predicting parent-level labels, which could be either positive or negative. Implementation details used for training can be found in section \ref{section:implementation}.

\section{Experiments}

We propose an entity and relation extraction task for radiology reports that can be developed using our development dataset and tested using our test dataset. Like the original RadGraph dataset, our dataset processes each radiology report into a sequence of space-delimited tokens, where punctuation like commas and semicolons have been separated from words to support entity recognition. For each report, we provide annotations identifying the type and span of each entity as well as relations between entities. 

\subsection{Evaluation}

\subsubsection{Evaluation Models}

We compare two entity and relation extraction model baselines against HGIE for entity and relation extraction tasks on RadGraph2. The first baseline uses the DyGIE++ framework \citep{wadden2019entity}, which was able to achieve state-of-the-art performance on the original RadGraph by jointly extracting entities and relations. Our second baseline uses the Princeton University Relation Extraction model (PURE) by Zhong et al. \citep{Zhong2021AFE}, which achieved state-of-the-art at the time of relation extraction using a pipeline approach that decomposes entity and relation extraction into different subtasks. We pit these baselines against HGIE which, as detailed above, builds upon the DyGIE++ framework to take advantage of the relationship between entities.

\subsubsection{Evaluation Method}

We evaluate our models on the entity and relation extraction tasks for radiology reports, using the training and validation partitions for training and the test partition for the final evaluation. Each of the input radiology reports is transformed into a sequence of space-delimited tokens with punctuation marks being considered as separate tokens. Every report is also associated with annotations that provide information about the spans and types for all entities and relations occurring in the report. 

In order to assess HGIE's performance in the medical sphere we evaluate all models on both RadGraph2 and the original RadGraph. With different definitions for relations, the introduction of new entities, and a vastly different entity hierarchy, both datasets are considerably different and thus allow for a comprehensive evaluation of our approach.

It is also important to determine how effective RadGraph2 is at performing tasks the original RadGraph dataset was capable of. We evaluate our benchmark model on the intersection of the test partitions of both datasets; that is, on the entities and relations that are the same between the two datasets. The benchmark model trained on RadGraph2 should ideally perform similarly to the one trained on RadGraph set when evaluated on the special set.

We report results on the MIMIC-CXR and CheXpert test partitions separately across all experiments. For more details on these partitions, refer to Section \ref{section:dataoverview}.

\subsubsection{Implementation Details}
\label{section:implementation}

We use BERT \citep{devlin2018bert} for all approaches. For DyGIE++, we use a learning rate of 5e-5 with a batch size of 1 for BERT and a learning rate of 1e-3 for task-specific layers, consistent with the approach used by Wadden et al. These same parameters are used for the first training phase of HGIE, while the fine-tuning phase uses a learning rate of 8e-6 with a batch size of 1 for BERT and a learning rate of 2e-4 for task-specific layers, which is consistent with the finetuning-approach used by \citep{xu-etal-2021-gradual}.

For PURE entity extraction, we use a learning rate of 1e-5 with a batch size of 16 for BERT, a span length of 3, and a learning rate of 5e-5 for task-specific layers. For PURE relation extraction, we use a learning rate of 2e-5, a context window of 50, and a batch size of 16 for BERT. These values are derived from \cite{radgraph-jain}.

For each of our approaches, in addition to using BERT weight initializations, we use weight initializations from five different biomedical pre-trained models: BioBERT \citep{lee2020biobert}, ClinicalBERT \citep{alsentzer2019publicly}, PubMedBERT \citep{gu2021domain}, BlueBERT \citep{peng2019transfer}, and RadBERT \citep{yan2022radbert}.

\subsubsection{Evaluation Metrics}
We report both micro and macro F1 for joint entity recognition and relation extraction for all models. A predicted entity is considered correct if both the predicted span boundaries and predicted entity type are correct. A predicted relation is considered correct if the pair of entities are correct, including both span boundaries and entity type, and the relation type is correct. As the performance on relation extraction also depends on the performance on the entity recognition (as the predicted relations cannot possibly be correct if the underlying entities are wrong), we focus our evaluation on this metric, as it provides a more holistic view of the performance of the model. Results on entity recognition can be found in Appendix \ref{appendix:ograd}. We separately report the results on the test data originating from the MIMIC-CXR and CheXpert.

\section{Results}

\begin{table}[!t]
     \caption{Relation extraction performance on RadGraph2. Best-performing values are bolded.}
     \centering
     \begin{center}
     \resizebox{0.7\linewidth}{!}{%
     \begin{tabular}{ c c c c c c } 
        \toprule
        \multirow{2}{*}{\textbf{Model}}  &
        \multicolumn{2}{c}{\textbf{MIMIC-CXR Relations}} & \multicolumn{2}{c}{\textbf{CheXpert Relations}} \\ 
        \cmidrule(r){2-3}
        \cmidrule(r){4-5}
       & \textbf{F1 micro} & \textbf{F1 macro} & \textbf{F1 micro} & \textbf{F1 macro} \\ 
        \toprule
        \multicolumn{1}{l}{\textit{DyGIE++}} \\
        BERT Base & 0.848 & 0.823 & 0.703 & 0.673 \\
        BioBERT & 0.850 & 0.815 & 0.714 & 0.689 \\
        ClinicalBERT & 0.847 & 0.783 & 0.703 & 0.642 \\
        PubMedBERT & 0.853 & 0.824 & 0.719 & 0.692 \\
        BlueBERT & 0.820 & 0.783 & 0.651 & 0.642 \\
        RadBERT & 0.851 & 0.832 & 0.723 & 0.693 \\
        \bottomrule
        \multicolumn{1}{l}{\textit{PURE}} \\
        BERT Base & 0.851 & 0.833 & 0.709 & 0.676 \\
        BioBERT & 0.850 & 0.812 & 0.715 & 0.691 \\
        ClinicalBERT & 0.841 & 0.785 & 0.708 & 0.649 \\
        PubMedBERT & 0.854 & 0.841 & 0.718 & 0.693 \\
        BlueBERT & 0.817 & 0.776 & 0.646 & 0.639 \\
        RadBERT & 0.855 & 0.843 & 0.718 & 0.698 \\
        \bottomrule
        \multicolumn{1}{l}{\textit{HGIE (Ours)}} \\
        BERT Base & 0.841 & 0.801 & 0.701 & 0.672 \\
        BioBERT & 0.853 & 0.838 & 0.727 & \textbf{0.718} \\
        ClinicalBERT & 0.841 & 0.815 & 0.715 & 0.691 \\
        PubMedBERT & 0.874 & 0.857 & 0.732 & 0.707 \\
        BlueBERT & 0.828 & 0.785 & 0.675 & 0.654 \\
        RadBERT & \textbf{0.879} & \textbf{0.862} & \textbf{0.739} & 0.714 \\
        \bottomrule
     \end{tabular}%
     }
     \end{center}
     \label{table:comp}
\end{table}

\paragraph{Taking advantage of hierarchy leads to improvements in performance.} We evaluate HGIE against our baseline models on both the MIMIC-CXR and CheXpert test sets. When comparing approaches, we use the strict relation extraction metric defined above as the primary end-to-end approach metric, as it uses both predicted entities and relations in its computation. Results are shown in Table \ref{table:comp}. HGIE is the best-performing model on this dataset, with the model outperforming its counterparts on both test sets. For most initializations, HGIE beats DyGIE++ and PURE in both F1 micro and macro on both test sets. Across all initializations, HGIE outperforms DyGIE++ on both test sets. HGIE and DyGIE++ possess the same model architectures, only differing in how they are trained, indicating that utilizing the entity taxonomy during training leads to better performance on RadGraph2. Appendix \ref{appendix:ograd} shows similar results for the original RadGraph dataset.

\begin{table}[!t]
     \centering
     \caption{Comparison of the entity extraction performance on the best performing RadGraph2 and RadGraph models on a common set of entities. Best-performing values are bolded.}
     \begin{center}
     \resizebox{0.95\linewidth}{!}{%
    \begin{tabular}{@{}ccccccc@{}}
\toprule
\multirow{2}{*}{Best Model} & \multirow{2}{*}{Anatomy} & Observation:       & Observation:   & Observation:      & \multirow{2}{*}{F1 micro} & \multirow{2}{*}{F1 macro} \\
                            &                          & Definitely Present & Uncertain      & Definitely Absent &                           &                           \\ \midrule
                            & \multicolumn{6}{c}{MIMIC-CXR}                                                                                                              \\ \cmidrule(l){2-7} 
RadGraph                    & 0.968 & 0.922 & 0.700 & \textbf{0.952} & 0.940 & 0.886                     \\
RadGraph2                   & \textbf{0.983} & \textbf{0.963} & \textbf{0.798} & 0.813 & \textbf{0.943} & \textbf{0.891} \\ \midrule
                            & \multicolumn{6}{c}{CheXpert}                                                                                                               \\ \cmidrule(l){2-7} 
RadGraph                    & \textbf{0.941} & 0.884 & 0.714 & \textbf{0.910} & \textbf{0.905} & 0.862            \\
RadGraph2                   & 0.880 & \textbf{0.939} & \textbf{0.812} & 0.823 & 0.894 & \textbf{0.864 }          \\ \bottomrule
\end{tabular}%
     }
     \end{center}
     \label{table:head-to-head-entities}
\end{table}

\begin{table}[!t]
     \centering
     \caption{Comparison of relation extraction performance on the best performing RadGraph2 and RadGraph models on a common set of entities and relations. Best-performing values are bolded.}
     \begin{center}
     \resizebox{0.7\linewidth}{!}{%
    \begin{tabular}{@{}cccccc@{}}
\toprule
Best Model & Modify         & Located At     & Suggestive Of  & F1 micro       & F1 macro       \\ \midrule
           & \multicolumn{5}{c}{MIMIC-CXR}                                                      \\ \cmidrule(l){2-6} 
RadGraph   & 0.804          & \textbf{0.861} & 0.685          & 0.823          & 0.783          \\
RadGraph2  & \textbf{0.906} & 0.826          & \textbf{0.838} & \textbf{0.874} & \textbf{0.857} \\ \midrule
           & \multicolumn{5}{c}{CheXpert}                                                       \\ \cmidrule(l){2-6} 
RadGraph   & 0.709          & \textbf{0.779}          & 0.588          &  0.725         & 0.692          \\
RadGraph2  & \textbf{0.777} & 0.712                   & \textbf{0.730} & \textbf{0.746} & \textbf{0.740} \\ \bottomrule
\end{tabular}%
     }
     \end{center}
     \label{table:head-to-head-rel}
\end{table}

\paragraph{RadGraph2 models perform at least as well as original RadGraph models on the same tasks.} We specifically evaluate the performance of the best-performing RadGraph2 model alongside the best-performing original RadGraph model on both entity recognition and relation extraction. These models are cross-compared between datasets by creating an intersection dataset. This is created by pruning entities and relations that are unique to RadGraph2. Results can be found in Tables \ref{table:head-to-head-entities} and \ref{table:head-to-head-rel}. 

Both benchmark models perform similarly on the entity extraction task, with the RadGraph2 model doing better for 2 out of 4 entity types on both MIMIC-CXR and CheXpert. The RadGraph2 model has higher performance on the more difficult relation extraction task, doing better for 3 out of 4 relation types on both MIMIC-CXR and CheXpert.

\begin{table}[hbt!]
     \caption{Effect of Entity Taxonomy Tree Depth on HGIE Performance. Best values are bolded.}
     \centering
     \begin{center}
     \resizebox{0.7\linewidth}{!}{%
     \begin{tabular}{ c c c c c} 
        \toprule
        \multirow{2}{*}{\textbf{Model}}  &
        \multicolumn{2}{c}{\textbf{MIMIC-CXR Relations}} & \multicolumn{2}{c}{\textbf{CheXpert Relations}} \\ 
        \cmidrule(r){2-3}
        \cmidrule(r){4-5}
       & \textbf{F1 micro} & \textbf{F1 macro} & \textbf{F1 micro} & \textbf{F1 macro}\\ 
        \toprule
        \multicolumn{1}{l}{\textit{Depth 3}} \\
        BERT Base & 0.841 & 0.801 & 0.701 & 0.672 \\
        BioBERT & 0.853 & 0.838 & 0.727 & \textbf{0.718} \\
        ClinicalBERT & 0.841 & 0.815 & 0.715 & 0.691 \\
        PubMedBERT & 0.874 & 0.857 & 0.732 & 0.707 \\
        BlueBERT & 0.828 & 0.785 & 0.675 & 0.654 \\
        RadBERT & 0.879 & 0.862 & 0.739 & 0.714 \\
        \bottomrule
        \multicolumn{1}{l}{\textit{Depth 2}} \\
        BERT Base & 0.841 & 0.801 & 0.702 & 0.660 \\
        BioBERT & 0.856 & 0.840 & 0.726 & 0.711 \\
        ClinicalBERT & 0.852 & 0.833 & 0.722 & 0.717 \\
        PubMedBERT & 0.881 & \textbf{0.862} & 0.741 & 0.716\\
        BlueBERT & 0.824 & 0.781 & 0.685 & 0.658 \\
        RadBERT & \textbf{0.883} & 0.861 & \textbf{0.742} & 0.717 \\
        \bottomrule
     \end{tabular}%
     }
     \end{center}
\end{table}

\paragraph{Modifying taxonomy structure can lead to improvements in performance.} So far, we have treated the structure of the entity taxonomy to be fixed. However, there are two different ways to categorize the entities, which can be defined based on the maximum depth of the taxonomy tree. Here, we seek to determine the effects of depth on the final model performance. We pit the depth 3 tree shown in Figures 1, 2, and 3 against a depth 2 tree where all change-related entities are children of the generic change node. Both models have similar performance to each other across model weight initializations. Nonetheless, the DGIE model trained under a taxonomy with a maximum depth of 2 tends to outperform the model with a taxonomy depth of 3. This is likely due to the small number of labeled entities for certain leaf nodes, like \texttt{CHAN-CON-RES} which only has a single example in the joint test set, adding noise during the conditional training phase. These results indicate that the structure of the tree in low data label environments should be considered akin to a tunable hyperparameter, with various depths evaluated in order to obtain optimal performance.

\section{Limitations}
Our study has several notable limitations. First, we only evaluate the HGIE model on radiology report datasets despite it being designed to generalize to any information extraction task where entities form a taxonomy. We do expect that the model training methodology proposed will work for other domains to solve similar problems, and strongly encourage research into further applications. Second, while care has been taken to make our RadGraph2 information schema as clear as possible, there are still certain cases in which the labeling can be ambiguous. For example, a radiologist may describe a change with a certain degree of uncertainty, which cannot be directly modeled by our schema. Third, the reports in our data are limited to chest X-ray radiology notes from MIMIC-CXR and CheXpert, though it is important to note that the RadGraph2 schema is designed such that it can annotate radiology reports in general. Fourth, the reports for both the MIMIC-CXR and CheXpert datasets are collected only from two United States hospitals (Beth Israel Deaconess Medical Center in Boston, MA, and Stanford Hospital in Stanford, CA, respectively). The distribution of the data may thus be substantially different from other hospitals and countries and the users planning to make use of the RadGraph2 dataset are urged to seriously consider the risks associated with domain shift.

Some ethical considerations associated with this work stem from our usage of medical reports from MIMIC-CXR and CheXpert. As in any situation in which real-world patient medical information is used in research, great care must be exercised in order not to compromise the privacy and rights of the human subjects from which the data has been collected. In our case, the risks associated with the usage of health information are greatly mitigated by the fact that the data has been carefully de-identified. Due to this fact, our work does not constitute human subject research and is exempt from a review by an institutional review board (IRB). Nevertheless, we still recognize that the data is potentially sensitive and thus take great care to use it responsibly and in accordance with its license.



\section{Conclusion}

We introduce RadGraph2, a new dataset of 800 chest X-ray expert annotations and an additional 220,913 annotated via inference. Radgraph2 combines the schema from the previous version of RadGraph with additional entities focused on characterizing priors along with clinically relevant contextual information in radiology reports. We characterize the entities via a hierarchy that follows the natural relationships between various types of entities. Via a modification of an information extraction model that has shown great promise in the medical sphere, we show that taking advantage of this hierarchy during training can lead to an improved joint entity and relation extraction performance.

We believe that our work paves the way for the development of systems capable of automatically characterizing disease progression in patients over time, as well as novel information extraction models leveraging the natural hierarchy of the labels in their target domain.


\bibliography{ref}

\newpage
\appendix
\section{Performance on Original RadGraph}
\label{appendix:ograd}

As discussed in the main manuscript, we wish to evaluate how well RadGraph2 encompasses the original dataset. To this end, we perform a head-to-head evaluation of the best models trained on each dataset and evaluate their performance on the intersection between their test sets.

To do this, we evaluate our models using the relation extraction task on the original RadGraph dataset. We constructed an entity taxonomy tree where observation-based entities were children of a generalized observation parent node and had a depth of 2. As \texttt{ANAT-DP} is the only type of anatomy node, we set it to be a leaf node at depth 1. 

Except for F1-macro on the MIMI-CXR test set with a PubMedBERT initialization, HGIE outperforms both baselines across all initializations. The best-performing model on this dataset was HGIE using a PubMedBERT initialization. Like on the RadGraph2 dataset, model weight initialization has a significant effect on final model performance but the deltas are slightly less pronounced on this dataset. F1 deltas for different initializations are as high as 0.040 for HGIE compared to 0.026 for PURE and 0.031 for DyGIE++.

DyGIE++ achieves an F1-micro of 0.826 and F1-macro of 0.787 for MIMIC-CXR Relations and achieves an F1-micro of 0.726 and F1-macro of 0.694 for CheXpert Relations. HGIE achieves an F1-micro of 0.852 and F1-macro of 0.791 for MIMIC-CXR Relations and achieves an F1-micro of 0.744 and F1-macro of 0.723 for CheXpert Relations. 

\begin{table}[ht]
     \caption{Relation extraction performance on RadGraph 1.0. Best-performing values are bolded. As was the case for RadGraph2, the best combination is HGIE with a RadBERT initialization.}
     \centering
     \begin{center}
     \resizebox{0.6\linewidth}{!}{%
     \begin{tabular}{ c c c c c c } 
        \toprule
        \multirow{2}{*}{\textbf{Model}}  & \multicolumn{2}{c}{\textbf{MIMIC-CXR Relations}} & \multicolumn{2}{c}{\textbf{CheXpert Relations}} \\ 
        \cmidrule(r){2-3}
        \cmidrule(r){4-5}
       & \textbf{F1 micro} & \textbf{F1 macro} & \textbf{F1 micro} & \textbf{F1 macro} \\ 
        \toprule
        \multicolumn{1}{l}{\textit{DyGIE++}} \\
        BERT Base & 0.805 & 0.752 & 0.712 & 0.688 \\
        BioBERT & 0.801 & 0.731 & 0.701 & 0.668 \\
        ClinicalBERT & 0.806 & 0.739 & 0.701 & 0.672 \\
        PubMedBERT & 0.823 & 0.783 & 0.725 & 0.692 \\
        BlueBERT & 0.803 & 0.712 & 0.705 & 0.664 \\
        RadBERT & 0.826 & \textbf{0.787} & 0.726 & 0.694 \\
        \bottomrule
        \multicolumn{1}{l}{\textit{PURE}} \\
        BERT Base & 0.805 & 0.731 & 0.722 & 0.648 \\
        BioBERT & 0.806 & 0.757 & 0.721 & 0.654 \\
        ClinicalBERT & 0.809 & 0.746 & 0.728 & 0.664 \\
        PubMedBERT & 0.812 & 0.745 & 0.729 & 0.679 \\
        BlueBERT & 0.818 & 0.738 & 0.699 & 0.655 \\
        RadBERT & 0.814 & 0.746 & 0.723 & 0.675 \\
        \bottomrule
        \multicolumn{1}{l}{\textit{HGIE (Ours)}} \\
        BERT Base & 0.809 & 0.752 & 0.719 & 0.699 \\
        BioBERT & 0.841 & 0.773 & 0.739 & \textbf{0.730} \\
        ClinicalBERT & 0.842 & 0.774 & 0.745 & 0.729 \\
        PubMedBERT & 0.849 & 0.779 & \textbf{0.752} & 0.726 \\
        BlueBERT & 0.838 & 0.740 & 0.730 & 0.699 \\
        RadBERT & \textbf{0.852} & 0.791 & 0.744 & 0.723 \\
        \bottomrule
     \end{tabular}%
     }
     \end{center}
\end{table}

\section{Annotation Instructions}

In this section, we provide the annotation instructions that were given to all annotators on the project.

\subsection{Datasaur Platform Instructions}

For this task, please label the reports assigned to you on datasaur.ai (a labeling platform). In addition to features shown in the video (some of which are not relevant, e.g. the ML-assisted labeling), you can also use the list of files under extensions on the right to navigate between the different batches of reports assigned to you and to mark these as complete once you have finished checking and modifying them.

Before you start labeling, please make sure that you select the correct set of labels for the arrow relations. You can do this by clicking on the dropdown menu marked with an arrow (in the top left part of the labeling interface) and selecting the second option from the top.

Please make sure that you use the dropdown on the top left rather than the one on the right side of the interface. You can check that you applied the correct label set by clicking on any arrow within a report — if the label set is correct, the available options should be located\textunderscore at, modify, and suggestive\textunderscore of. 

\subsection{General Task Information}
The labeling task is to correct any mistakes or deficiencies in the existing RadGraph labels. The labels consist of entities (Change, Anatomy, and Observation) and relations between them (suggestive\textunderscore of, located\textunderscore at and modify) marked in each radiology report. Most of these entities and relations are already in place and are shown in the Datasaur interface, but some of them may be missing, mislabelled or improperly selected. Your main task is to look for any comparisons to the prior radiology examinations mentioned in the reports and ensure that they are correctly marked using the appropriate Change entities (described below) and relations connected to them. Thus, you may need to change the types of certain entities or relations, add new ones, or (in some cases) remove existing ones. The Anatomy and Observation entities and the relations between them should be substantially more complete and reliable, so you are not expected to actively check them (provided that they do not mark comparison to a prior, in which case they should be converted into a Change entity), but please feel free to amend or modify them if you spot any obvious mistakes.

Update: Please focus on correctly marking the change entities and the relations associated with them and try not to make extensive changes to the entities and relations not related to changes. Please see the “Datasaur annotator feedback” document for details. (Note: The referred-to document includes names and provides personalized feedback to the annotators. In order to not break anonymity, these details will not be included.)

\subsection{Entities}

\begin{outline}
\1 Change (CHAN): an expression explicitly describing change or lack of change compared to prior or indicating that comparison to prior has occurred. Examples of change entities include “stable”, “change”, “unchanged”, “new”, “larger”, “comparison”. There are several different types of change entities, which we describe below along with examples. Each example shows the text span(s) marked with the given entity type, highlighted in bold. We also give the outgoing relations from the change entity (see below for information regarding relations).

    \2 CHAN-NC indicates a lack of change since the prior study. Note that this does not cover cases in which a finding is reported without explicit information of whether it is different from the prior or not — in these cases, the change entity types should not be used at all (instead, the appropriate OBS- or ANAT- entities should be marked).
        \3 In comparison with the earlier study of this date, there is essentially no change in the appearance of the enteric tube.
            \4 change modifies tube
        \3 Moderately severe bibasilar atelectasis persists.
            \4 persists modifies atelectasis
        \3 Cardiomediastinal silhouette is unchanged.
            \4 unchanged modifies silhouette

    \2 CHAN-IMP indicates an improvement in a certain aspect of the patient’s clinical state compared to the prior. If a certain adverse medical condition of the patient was described as being completely resolved, use CHAN-CON-RES instead.
        \3 Left basilar opacity has nearly resolved in the right lower lobe opacity has improved.
            \4 resolved modifies opacity (first occurrence) — note that the condition has not completely resolved, so the change is only considered to be an improvement
            \4 improved modifies opacity (second occurrence)
        \3 Compared to the most recent study, there is improvement in the mild pulmonary edema and a decrease in the small left pleural effusion.
            \4 Improvement modifies edema
            \4 decrease modifies effusion

    \2 CHAN-WOR indicates a worsening in a certain aspect of the patient’s clinical state compared to the prior. If a certain adverse medical condition of the patient was described as being completely new, use CHAN-CON-AP instead.
        \3 Mild - to - moderate diffuse pulmonary edema is slightly worse.
            \4 worse modifies edema
        \3 Moderate right pleural effusion has increased since...
            \4 increased modifies effusion

\2 CHAN-CON-AP indicates that, compared to the previous report(s), a new adverse medical condition has been observed in the given patient.
\3 There is also a new left basilar opacity blunting the lateral costophrenic angle (...)
\4 new modifies opacity

\2 CHAN-CON-RES indicates that compared to the previous report(s), a certain medical condition previously observed in the patient has completely resolved
\3 Indistinct superior segment left lower lobe opacities have resolved.
\4 resolved modifies opacities

\2 CHAN-DEV-AP indicates that, compared to the previous report(s), the patient has been fitted with a new medical device or tool (e.g. intubated, catheterized, …)
\3 The patient has received the new feeding tube.
\4 new modifies tube

\2 CHAN-DEV-PLACE indicates that the position of a medical device in the body of a patient changed compared to prior studies.
\3 Left pleural drain has been pulled in a slightly higher position.
\4 higher modifies position
\3 Left pleural drain has been advanced to the left apex.
\4 Advanced modifies drain

\2 CHAN-DEV-DISA indicates that, compared to the previous report(s), a medical device or tool was detached or removed from the patient.
\3 In the interval, the patient has been extubated (...)
\4 Extubated modifies patient
\3 The nasogastric tube has been removed.
\4 removed modifies tube

\2 The change entity types are used solely for spans of tokens indicative of change (or no change) in the report texts, not for entities representing medical conditions, devices, etc. related to the change. Instead, the entity denoting change will typically be attached to the other relevant entities in the report via a relation. This is demonstrated in the above examples.

\2 Anatomy (ANAT-DP): an anatomical body part that occurs in the radiology report. Examples of anatomy entities include “lung”, “left lower lobe of the lung” (multiple entities), or “aortic arch” (multiple entities).
\3 Anatomy vs. Anatomy Modifier: in this schema, all anatomy modifiers are annotated as anatomy entities. In the case that anatomy modifies the scope or degree of a second anatomy, a “modify” relation is added to denote the modification relationship between the two entities. See below for a definition and example of the “modify” relation. So, "left lung" would be two anatomy entities, where “left” modifies “lung”.

\2 Observation (OBS): an observation made from the images and associated with visual features, identifiable pathophysiologic processes or diagnostic disease classifications. Examples of multiple observation entities include “airspace opacity”, “mass”, “bilateral pleural effusion” or “pneumonia”. 
\3 Observation vs. Observation Modifier: in this schema, all observation modifiers are annotated as observation entities. In the case that an observation modifies the scope or degree of a second observation, a “modify” relation is added to denote the modification relationship between the two entities. See below for a definition and example of the “modify” relation.

\2 Note that each observation entity is associated with an uncertainty attribute measuring the uncertainty level of an observation entity or anatomy entity. Each uncertainty attribute can have one of three values: definitely present (OBS-DP), uncertain (OBS-U), or definitely absent (OBS-DA). 
\3 When you label an observation, you will pick from the following options: Observation::definitely present (OBS-DP), Observation::uncertain (OBS-U), or Observation::definitely absent (OBS-DA).
\3 In some cases text spans for Observations are not continuous.  In that case, each span should be labeled as observation with a “modify” relation between them.

\end{outline}

\subsection{Relations}
\begin{outline}

\1 Relations are directed arrows from one entity to another that are used to describe a relationship between two entities. We define the following three types of relations, in the form of “relation\textunderscore type (entity \textunderscore type, entity\textunderscore type)”:

\2 suggestive\textunderscore of (observation, observation) or (change, observation) or (observation, change): a relation between two observation entities indicating that the status of the second observation is inferred from that of the first observation or indicating that change is derived from an observation or indicating that an observation is derived from a change.

\2 located\textunderscore at (observation, anatomy): a relation between an observation and an anatomy entity indicating that the observation is related to the anatomy. While located\textunderscore at often refers to location, it can also be used to describe other relations between an observation and an anatomy. For example, in the sentence, “heart is normal”, “normal” is an observation, “heart” is an anatomy, and “located\textunderscore at” is a relation from “normal” to “heart.”

\2 modify (observation, observation) or (anatomy, anatomy) or (change, any) or (observation, change): a relation between two observation entities indicating that the first observation modifies the scope of or quantifies the degree of the second observation. This relation is often added when the first observation entity is an “observation modifier” that modifies the second observation. The same logic applies to an anatomy entity modifying another anatomy entity. Additionally, the modify relationship is commonly used to connect change entities to other entities, as demonstrated in the examples above. In some cases, the modify relation can also be used to connect an observation entity to a change entity, as shown in some of the examples in this document.

\end{outline}

\subsection{Change Entities Annotation Conventions}

\begin{outline}
\1 In some cases, there might be multiple ways in which to mark the change entities in the given report. In order to improve the consistency of the data, we adopted several conventions:
\2 Sometimes, there might conceivably be multiple spans of words indicating a certain change category. We generally strive to select the minimum possible span related to the description of the change type in the note text. For example, if the note says “Biapical scarring is again seen”, we mark the entity “again” as CHAN-NC rather than using an entity with a longer span “again seen”. Similarly, if the note says “In comparison with the earlier study of this date, there is essentially no change in the appearance of the enteric tube”, we only mark the entity “change” as CHAN-NC instead of marking an entity “no change” (the model using the annotations will still consider the wider context, so “no” does not need to be annotated at all).
\2 We almost exclusively use the “modifies” relation when connecting the change entities with the rest of the graph. If the change entity is connected to the rest of the graph using a “located\textunderscore at” relation, please consider changing it to “modifies”. Additionally, the change entity should only be connected to the main entity/entities that is/are related to the change.
\2 If the report indicates uncertainty with regards to the presence of a change, we will usually attempt to ascertain whether the presence of the change was perceived to be likely by the author of the note. If yes, we mark the change, otherwise, we annotate these entities using the “OBS-” markers from the origin=al RadGraph (version 1)
\2 We mark all mentions of a change in a report, even if some of the mentions are duplicate (i.e. they repeat information about an identical change). Additionally, the reports will often contain multiple changes of different types (e.g. if certain aspects of the patient’s medical state improved and others worsened).

\end{outline}

\subsection{Suggested Annotation Process}
\begin{outline}
\1 (Note: These original instructions for annotating the reports are included for information only, you are only expected to verify and correct the existing labels)

\2 Here is a suggested process for each sentence or sentence group:
\3 Find the main anatomy (if present), and label as anatomy
\3 Find any anatomic modifiers. Label these modifiers as anatomy and link each to main anatomy using “modify”.
\3 Find all observations and label them as observations
\3 Link all observations to the appropriate anatomy using “located\textunderscore at” link.
\3 Find any observation modifiers and label as observations with modifier link to observations they modify
\3 Take note of any mentions of change (or no change) and label them appropriately
\3 Link the change entities to the rest of the graph as appropriate
\2 Here is how this process would be applied to the following example: “The heart is top normal in size, though this is stable”.
\3 Main anatomy is heart.
\3 size is an attribute of heart, so it is also labeled as anatomy with “modify” relation to heart.
\3 normal is an observations with “located\textunderscore at” relation to the main anatomy, heart. Note that top is an observation with “modify” relation to normal.
\3 stable is a change (in particular, CHAN-NC) with “modifies” relation to the main aspect which is stable, i.e. normal

\end{outline}

\subsection{Other Specific Annotation Rules}
\begin{outline}
\1 Please follow these additional rules.
\2 The main anatomy may be an adjective or may be part of a compound term.  For pleural effusion, pleural is anatomy and effusion is observation.  A common discrepancy was labeling the entire phrase as the observation.
\2 Words such as size, volume, wall of, silhouette, structure, length, near, below, are typically attributes or modifiers of anatomy: e.g., lung volume, heart size, cardiac silhouette, osseous structure, below the diaphragm, near the apex.
\2 By convention, compound words, such as cardiomegaly, which means an enlarged heart and includes both anatomy and observation semantics, should be labeled as an observation. In the same way, pneumothorax or cirrhosis are observations with the anatomic location only implied by the term itself.
\2 Terms like left, right, and bilateral should be considered anatomy modifiers and labeled as anatomy. For example, in left kidney, both left and kidney are anatomy. For bilateral pleural effusions, bilateral is a modifier for the primary anatomy (pleural).
\2 When an anatomy and the modifier(s) are next to each other, identify a primary term and the modifier(s) separately for consistency even if they are adjacent.  For example, bilateral pleural, should be labeled as two observations, with bilateral modifying pleural. This creates consistency with the case where the two tokens are distant in text, such as bilateral small effusions or large effusions which are now bilateral.
\2 A word in a sentence can sometimes modify a word in a previous sentence. For example, in the sentences, “The tube extends to the stomach. The tip is near the GE junction”, the word “tip” should have a modifier link to the word “tube” in the first sentence.
\2 Whenever there is any degree of uncertainty, mark the anatomy or observation as uncertain. For the phrases “No signs of pneumonia” and “No evidence of pneumonia”, pneumonia is definitely absent. For the phrases “No definite signs of pneumonia,” “No clear signs of pneumonia,” and “No obvious signs of pneumonia,” pneumonia is uncertain due to the additional qualifier.

\end{outline}



\end{document}